%% file: affecteval.tex
\documentclass[sigconf]{acmart}

\renewcommand\footnotetextcopyrightpermission[1]{}
\AtBeginDocument{%
  }

\input{macro.tex}

\begin{document}

\title{\toolname: A Modular and Customizable Affective Computing Framework}

\author{Emily Zhou}
\email{emilyzho@usc.edu}
\orcid{0009-0009-5004-6588}
\affiliation{
  \institution{Computer Science\\University of Southern California}
  \city{Los Angeles}
  \state{CA}
  \country{USA}
}

\author{Khushboo Khatri}
\email{khatrik@usc.edu}
\orcid{0009-0001-3336-392X}
\affiliation{
  \institution{Electrical and Computer Engineering\\University of Southern California}
  \city{Los Angeles}
  \state{CA}
  \country{USA}
}

\author{Yixue Zhao}
\email{yzhao@isi.edu}
\orcid{0000-0003-3046-6621}
\affiliation{
  \institution{USC Information Sciences Institute\\University of Southern California}
  \city{Arlington}
  \state{VA}
  \country{USA}
}

\author{Bhaskar Krishnamachari}
\email{bkrishna@usc.edu}
\orcid{0000-0002-9994-9931}
\affiliation{
  \institution{Electrical and Computer Engineering\\University of Southern California}
  \city{Los Angeles}
  \state{CA}
  \country{USA}
}

\begin{abstract}
The field of affective computing focuses on recognizing, interpreting, and responding to human emotions, and has broad applications across education, child development, and human health and wellness. However, developing affective computing pipelines remains labor-intensive due to the lack of software frameworks that support multimodal, multi-domain emotion recognition applications. This often results in redundant effort when building pipelines for different applications. While recent frameworks attempt to address these challenges, they remain limited in reducing manual effort and ensuring cross-domain generalizability. We introduce \toolname, a modular and customizable framework to facilitate the development of affective computing pipelines while reducing the manual effort and duplicate work involved in developing such pipelines. We validate \toolname~ by replicating prior affective computing experiments, and we demonstrate that our framework reduces programming effort by up to 90\%, as measured by the reduction in raw lines of code.
\end{abstract}


\maketitle
\pagestyle{empty}

\input{sections/1_intro}
\input{sections/2_related_work}
\input{sections/3_framework_design}
\input{sections/4_system_evaluation}
\input{sections/5_discussion}
\input{sections/6_conclusion}

\bibliographystyle{ACM-Reference-Format}
\bibliography{reference}

\end{document}

%% file: macro.tex
\usepackage{setspace}
\usepackage{algorithmic}
\usepackage{graphicx}
\usepackage{textcomp}
\usepackage{xcolor}
\usepackage{soul}
\usepackage{comment}
\usepackage[linesnumbered,ruled,vlined]{algorithm2e}
\usepackage{balance}
\usepackage{microtype}
\usepackage{multirow}
\usepackage[normalem]{ulem}
\usepackage{soul}
\usepackage{lipsum}
\usepackage{wrapfig}
\usepackage{graphicx}
\usepackage{subcaption}
\usepackage{tabularray}
\usepackage{xargs}
\usepackage{stfloats}
\usepackage{float}
\usepackage{adjustbox}
\usepackage{listings}
\usepackage{caption} 

\definecolor{amethyst}{rgb}{0.6, 0.4, 0.8}

\usepackage[colorinlistoftodos, prependcaption, textsize=tiny]{todonotes}
\newcommandx{\khushboo}[2][1=]{\todo[linecolor=green, backgroundcolor=green!25, bordercolor=green, #1]{#2}}
\newcommand{\code}[1]{\texttt{#1}}

\newcommand{\toolname}{\textbf{AffectEval}}

\DeclareFixedFont{\ttb}{T1}{txtt}{bx}{n}{8} 
\DeclareFixedFont{\ttm}{T1}{txtt}{m}{n}{8}  

\definecolor{deepblue}{rgb}{0,0,0.5}
\definecolor{deepred}{rgb}{0.6,0,0}
\definecolor{deepgreen}{rgb}{0,0.5,0}
\newcommand\pythonstyle{\lstset{
    language=Python,
    basicstyle=\ttm,
    morekeywords={self},              
    keywordstyle=\ttb\color{deepblue},
    emph={MyClass,__init__},          
    emphstyle=\ttb\color{deepred},    
    stringstyle=\color{deepgreen},
    frame=tb,                         
    showstringspaces=false
}}
\newcommand\pythoninline[1]{{\pythonstyle\lstinline!#1!}}
\lstnewenvironment{python}[1][]{ \pythonstyle \lstset{#1}}{}

%% file: sections/1_intro.tex
\section{Introduction}
\label{sec:intro}
Affective computing is an interdisciplinary field of research for recognizing, interpreting, and responding to human affect with computational methodologies. While stimuli may vary, many affective computing pipelines share the same setup. \citet{metafera} outlines the key steps of signal acquisition, signal preprocessing, feature extraction and prediction as the key steps of the affective computing pipeline, which is illustrated in Figure \ref{fig:pipeline}. As a result, several software libraries and frameworks have been developed to support the creation of affective computing pipelines \cite{affecttoolbox, biosppy, covidentify, metafera, neurokit}. However, due to the broad applications of affective computing\textemdash which encompasses mental health (MH) disorders \cite{zucco2017, assabumrungrat2022, greene2016}, personalized education \cite{bota2019review}, and aging-related diseases such as Alzheimer's \cite{smith2021_affective_computing}\textemdash these software frameworks are not \textit{comprehensive} and have additional limitations that create high manual costs and prevent them from being reusable. A \textit{comprehensive} framework for affective computing should provide all the components necessary to implement a wide range of use cases across domains, signals, and methodologies. 

To guide the development of such a framework, we analyzed existing affective computing tools. Programming languages such as Python \cite{python} offer libraries such as scikit-learn \cite{scikit-learn}, pyHRV \cite{pyhrv}, and NeuroKit2 \cite{neurokit} that support parts of the affective computing pipeline. This includes signal preprocessing, feature extraction, and classification. Recent work has also introduced frameworks aimed at being more flexible \cite{metafera} and user-friendly \cite{affecttoolbox} to facilitate the development of affective computing pipelines. However, we identify the following main limitations: 

\begin{enumerate}
    \item Researchers typically need to create pipelines from scratch using various existing libraries and frameworks. This often requires extensive software experience, is time-consuming, and results in duplicated work across applications.
    \item Generally, affective computing pipelines are signal- and domain-specific; \textit{i.e.}, they only support a limited set of signals and applications. As a result, it is difficult to reuse code across pipelines or make them generalizable across datasets. 
    \item There is no standardized interface to compare the performance of individual components across different affective computing techniques. 
\end{enumerate}

To the best of our knowledge, there is no comprehensive framework for the development of end-to-end affective computing systems. To address the aforementioned limitations, such a framework should be \textit{modular}, \textit{customizable}, and \textit{flexible} to support multimodal signals, \textit{i.e.}, from multiple types of sensors and data streams, and multi-domain applications, such as general affect, stress, and depression detection. A framework consisting of individual modules, each responsible for a different component of the affective computing pipeline, provides an easily reusable structure. Users should also be able to customize modules as needed to create new pipelines or adapt existing ones for new applications. They can also compare the performance of different components in a fair playground by modifying their behaviors, such as using different signal preprocessing methods, feature extraction techniques, and classification models. 

We present \toolname: a modular and
customizable affective computing framework that generalizes to more signals and applications than existing frameworks. Our framework is designed under the object-oriented paradigm such that each component is responsible for a separate part of the affective computing pipeline. We program each component with default functionalities that can also be overwritten with custom methods, allowing researchers to easily set up basic end-to-end pipelines or create specialized applications as needed. We primarily focus on the use of time-series physiological signals for affective computing due to their extensive usage in emotion recognition and connection to mental states as suggested by the Somatic Marker Hypothesis \cite{damasio_somatic}, but \toolname~ can be easily extended to support other types of signals. We also establish a standard format for affective computing datasets in order to mitigate the amount of manual setup work required, which we detail in Section \ref{subsec:prereq}.

Additionally, we demonstrate the functionality of \toolname~ by reproducing the experiments of Schmidt et al. \cite{schmidt2018} and Zhou et al. \cite{zhou2023} using the Anxiety Phases Dataset (APD) \cite{senaratne2021} and Wearable Stress and Affect Detection dataset (WESAD) \cite{schmidt2018}. These datasets were chosen because they are well-established multimodal datasets for stress and affect detection. Both authors conducted different types of affect recognition tasks: Schmidt et al. performed binary and three-class affect classification on WESAD, while Zhou et al. performed binary stress detection on both APD and WESAD. These datasets and experiments are therefore suitable for demonstrating the ease-of-use and reusability of \toolname, as well as its ability to create signal- and domain-agnostic pipelines. \textit{Our \toolname-based pipeline's model performance largely matches or exceeds the performance achieved in previous work \cite{zhou2023, schmidt2018}, while using up to 90\% fewer lines of code.} We reuse the pipeline structure across our experiments and incorporate different preprocessing techniques, feature extraction methods, and models using \toolname's modular design.

In summary, this paper makes the following contributions:
\begin{itemize}
    \item We conduct an extensive literature review on existing affective frameworks to motivate our design of \toolname. 
    \item We develop \toolname, a comprehensive framework with pre-implemented behavior to facilitate affective computing research. We open-source the framework and artifacts to foster future research in this domain.
    \item We migrate existing work to \toolname, providing a starting point for researchers to compare affective computing works under a fair playground.  
    \item We reproduce existing work, quantifying the reduction of manual work \toolname~ achieves while meeting or exceeding their model performance metrics.
\end{itemize}

Section \ref{sec:related} provides an overview of existing methodologies, software libraries, and pipelines for affective computing. Section \ref{sec:design} describes the architecture of \toolname~ and outlines the steps necessary to implement an affective computing pipeline using our framework. Section \ref{sec:system_evaluation} describes our implementation of \toolname~ to replicate Schmidt et al. and Zhou et al's experiments. Section \ref{sec:discussion} discusses our findings. We conclude the paper in Section \ref{sec:conclusion}.

%% file: sections/2_related_work.tex
\input{tables_and_figures/related_work}

\section{Background and Related Work}
\label{sec:related}
In this section, we review the potential of affective computing in human-centered applications with a focus on mental healthcare, followed by an overview of existing software libraries and pipelines for affective computing. We identify their strengths, disadvantages, and common components and use them as guidelines for the development of \toolname.

\subsection{Affective Computing}
Affective computing has shown significant promise in mental health applications, particularly in the diagnosis, tracking, and treatment of various disorders. Recent studies have explored its use in a wide range of mental healthcare applications, especially depression \cite{zucco2017} and anxiety detection \cite{jiang2023}. Affective computing can also benefit the treatment and care of late-life mood and cognitive disorders, such as depression and Alzheimer's disease, via objective biomarkers such as eye movement and vocal features \cite{smith2021_affective_computing}.

The types of signals used in affect detection for health applications vary, each with its own advantages and limitations. Facial analysis and audio processing are popular due to their ease of collection and non-invasive nature \cite{liu2024affectivecomputinghealthcare}. They can be particularly useful in the detection of emotions such as stress or anxiety in clinical settings, but may be subject to cultural biases and environmental factors. Physiological signals, such as EEG, ECG, and skin conductance, offer a more objective measure of emotional states due to their direct connection to the autonomic nervous system \cite{Luneski2008-vt}. Recent works have used physiological signals to perform stress \cite{kuttala2023, giannakakis2022, schmidt2018, zhu2019}, and depression \cite{egger2019, zucco2017} detection.

\subsection{Investigation of Existing Affective Computing Pipelines}
Based on previous work, an affect recognition system can be broken down into the following main components: signal acquisition, signal pre-processing, feature extraction, feature selection, and classification. Table \ref{tab:related-work} provides an overview of related work and the components incorporated in each one. Next, we illustrate the tasks carried out by each component in the context of Schmidt et al.'s experiments \cite{schmidt2018}.

\begin{figure}[!htbp]
  \includegraphics[width=0.5\textwidth]{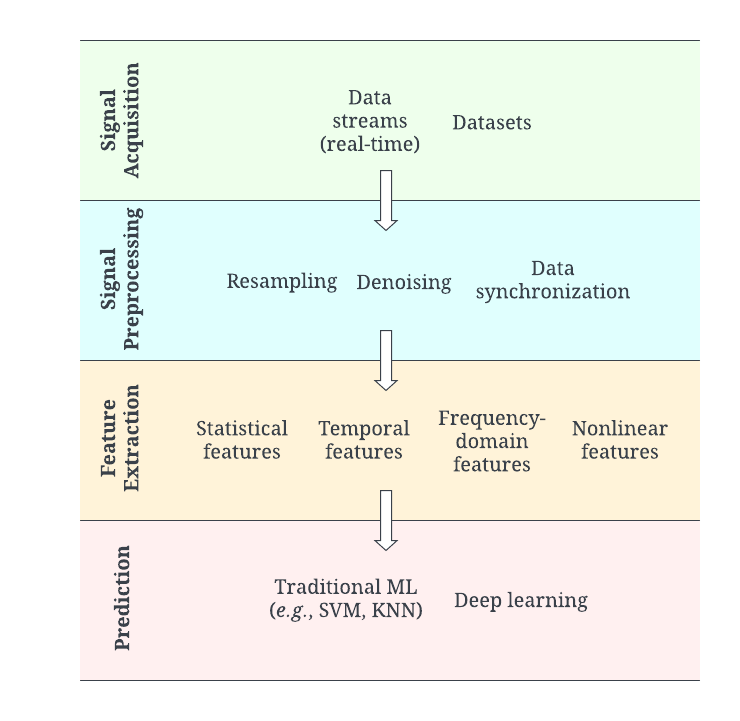}
  \caption{Key components of an affective computing pipeline, identified in \cite{metafera}, and examples of actions performed by each.}
  \label{fig:pipeline}
\end{figure}

\textbf{Signal acquisition} is the process of reading in signals to the affect recognition system. This can be from a local database or in real time from more data streams \cite{aranha2021, wagner2009}. In recent years, many benchmark datasets of unimodal and multimodal affect have been released \cite{wang2022} to support affective computing research. The methods of emotion elicitation and types of signals collected vary widely. For example, the Wearable Stress and Affect Detection (WESAD) dataset \cite{schmidt2018} contains blood volume pulse (BVP), electrocardiogram (ECG), electrodermal (EDA), electromyography (EMG), respiration (RESP), body temperature (TEMP), and three-axis acceleration (ACC) collected from neutral, stress, and amusement states. DEAP \cite{deap} contains electroencephalogram (EEG), peripheral physiological signals, and audiovisual recordings from individuals watching music videos. There have also been efforts to develop real-time emotion recognition systems; \cite{cheng_real_time} introduced a method for real-time negative affect detection from an ECG signal stream, while \cite{bertero_real_time} developed an interactive dialogue system to recognize users' emotion in real-time.

\textbf{Signal preprocessing} is necessary to remove noise and isolate frequency bands for feature extraction. \cite{schmidt2018} used various filters; for example, they applied a high-pass filter to remove the DC component from EMG signals, and a low-pass filter to denoise EDA signals. Other common preprocessing techniques are removing baseline wander and powerline noise from ECG signals. 

\textbf{Feature extraction} is performed in classical feature-based approaches to affect recognition \cite{bota2019review} using non-DL models, such as Support Vector Machine (SVM) \cite{svm} and Random Forest \cite{random_forest}. For examples, \cite{schmidt2018} performed manual feature extraction, obtaining statistical features (mean, median, standard deviation) and high-level physiological features. From ECG signals, heart rate, heart rate variability (HRV), and frequency-domain features were further extracted, and EDA signals were decomposed into their tonic and phasic components. Such high-level features have been found to be indicative of certain emotional states, demonstrating the necessity of performing feature extraction. For example, decreased HRV is associated with elevated anxiety \cite{kreibig2010autonomic}, and changes in the SCL may reflect changes in arousal \cite{braithwaite}. 

\textbf{Feature selection} is often necessary to reduce the dimensionality of the input features \cite{bota2019review}. While this step was not included in \cite{schmidt2018}, feature selection and dimensionality reduction can be performed to prevent over-fitting \cite{zhu2019}. Some commonly used feature selection methods include Principal Component Analysis (PCA) \cite{pca}, which reduces a feature set based on its uncorrelated \textit{principal components}, and Sequential Forward Selection (SFS) \cite{gelsama1994sfs}, which is an iterative approach that adds features that provide the most improvement in classifier performance.

\textbf{Classification} is the final step in the pipeline, consisting of training and evaluating the affect recognition model. Both traditional machine learning (ML) and deep learning (DL) methods can be used to perform binary and multi-class classification of affect \cite{bota2019review, saganowski2023review}. In particular, \cite{schmidt2018} used the Decision Tree (DT), Random Forest (RF), AdaBoost (AB), Linear Discriminant Analysis (LDA), and K-nearest neighbors (KNN) ML algorithms to perform binary stress classification and three-class classification of non-stress, stress, and amusement. Others have performed valence and arousal classification \cite{deap} based on Russell's circumplex model of affect \cite{circumplex_model}.

\subsection{Existing Software Libraries and Pipelines}
Existing software libraries and frameworks for affective computing generally support a few, but not all components of the pipeline. For example, Matlab \cite{matlab} offers many packages that provide functions for signal processing, analysis, and visualization. PyTorch \cite{pytorch} and Tensorflow \cite{tensorflow} are open-source frameworks to build machine learning models, but can require complex scripts that are time-consuming to set up. Other libraries focus on pre-processing methods for a small subset of signals, such as \textit{biosppy} \cite{biosppy} and \textit{neurokit} \cite{neurokit} for physiological signals, including BVP, ECG, EDA, EEG, and EMG, and \textit{heartpy} \cite{heartpy} for ECG and PPG signals.

While there exist more comprehensive tools that support more components of the pipeline, they are typically specific to one or very few domains of application. The AffectToolbox \cite{affecttoolbox} aims to facilitate the development of affective computing applications for research via a graphical user interface (GUI) framework, removing the need for extensive programming knowledge. It offers signal processing and deep learning methods for video and audio signals, and provides real-time analysis. While it achieves the goal of being accessible for a wider range of users across disciplines, it does not support the use of multimodal physiological signals.

The metaFERA \cite{metafera} framework provides building blocks to create domain-specific software frameworks for emotion recognition. It consists of independent and reusable modules, providing the high-level structure of an affective computing pipeline. However, it does not provide default functionalities or support deep learning models, requiring the user to implement their own methods. It is also written in Java, while the majority of libraries used in affective computing are available in Python. This could pose a problem for researchers in this domain, since they are less likely to be familiar with Java.

In general, existing software tools for affective computing  \textit{only support a small portion of the pipeline or subset of signals and have limited use cases}. As a result, there is a need for an affective computing framework that encompasses the full pipeline from signal acquisition to model evaluation, while being low-effort to implement and easily customizable.

%% file: tables_and_figures/related_work.tex
\definecolor{green}{rgb}{0.3, 0.60, 0.35}
\newcommand{\cmark}{{\color{green}\checkmark}}
\newcommand{\xmark}{{\color{red}\large{X}}}

\begin{table*}[ht]
\begin{adjustbox}{width=0.95\textwidth}
\begin{tabular*}{\textwidth}{ccccccccc}
\toprule
& \multirow{2}{*}{\raisebox{-2ex}{Paper}} & \multirow{2}{*}{\raisebox{-4ex}{\shortstack[c]{Multimodal\\Signals}}} & \multirow{2}{*}{\raisebox{-4ex}{\shortstack[c]{Multi-\\Domain}}} & \multicolumn{5}{c}{Affective Computing Pipeline Components} \\ \cmidrule{5-9}
& & & & \small \raisebox{1.5ex}{Preprocessing} & \small{\shortstack[c]{Feature\\Extraction}} & \small \small{\shortstack[c]{Feature\\Selection}} & \small \small{\raisebox{1.5ex}{Traditional ML}} & \small \small{\shortstack[c]{Deep\\Learning}} \\
\midrule 

\multirow{5}{*}{\shortstack[c]{\textbf{Affective}\\\textbf{Computing}\\\textbf{Experiments}}} & Zhu et al. \cite{zhu2019} & \cmark & \xmark & \cmark & \cmark & \cmark & \cmark & \cmark \\ 
& Khateeb et al. \cite{khateeb2021} & \cmark & \xmark & \cmark & \cmark & \xmark & \cmark & \xmark \\
& Koelstra et al. \cite{deap} & \cmark & \xmark & \cmark & \cmark & \xmark & \cmark & \xmark \\
& \small{Dominguez-Jimenez et al.} \cite{dominguez2020} & \cmark & \xmark & \cmark & \cmark & \cmark & \cmark & \xmark \\
& Ayata et al. \cite{ayata2017emotion} & \xmark & \xmark & \cmark & \cmark & \xmark & \cmark & \xmark \\
& Schmidt et al. \cite{schmidt2018} & \cmark & \cmark & \cmark & \cmark & \xmark & \cmark & \xmark \\
& Zhou et al. \cite{zhou2023} & \cmark & \xmark & \cmark & \cmark & \xmark & \cmark & \xmark \\

\midrule

\multirow{3}{*}{\textbf{Frameworks}} & Oliveira et al. \cite{metafera} & \cmark & \cmark & \cmark & \cmark & \cmark & \cmark & \xmark \\
& Mertes et al. \cite{affecttoolbox} & \cmark & \cmark & \cmark & \cmark & \xmark & \cmark & \cmark \\ 
& Wagner et al. \cite{wagner2009} & \cmark & \cmark & \cmark & \cmark & \xmark & \cmark & \xmark \\

\midrule

& \toolname & \cmark & \cmark & \cmark & \cmark & \cmark & \cmark & \cmark \\

\bottomrule

\end{tabular*}
\end{adjustbox}
\caption{Examples of affective computing experiments and frameworks containing the components of affective computing pipelines described in Oliveira et al. \cite{metafera}. \toolname~ is the first open-source, modular affective computing framework containing \textit{all} pipeline components.}
\label{tab:related-work}
\end{table*}

%% file: sections/3_framework_design.tex
\section{\toolname's Design}
\label{sec:design}
This section describes the design of \toolname~ and its features that address the aforementioned shortcomings of existing affective computing libraries and frameworks. \toolname~ is a meta-framework that enables the straightforward construction of end-to-end affective computing pipelines via building blocks defined using individual classes. \toolname~ has 3 key characteristics as discussed below, directly addressing existing works' limitations.

\begin{figure*}[!tbp]
    \includegraphics[width=\textwidth]{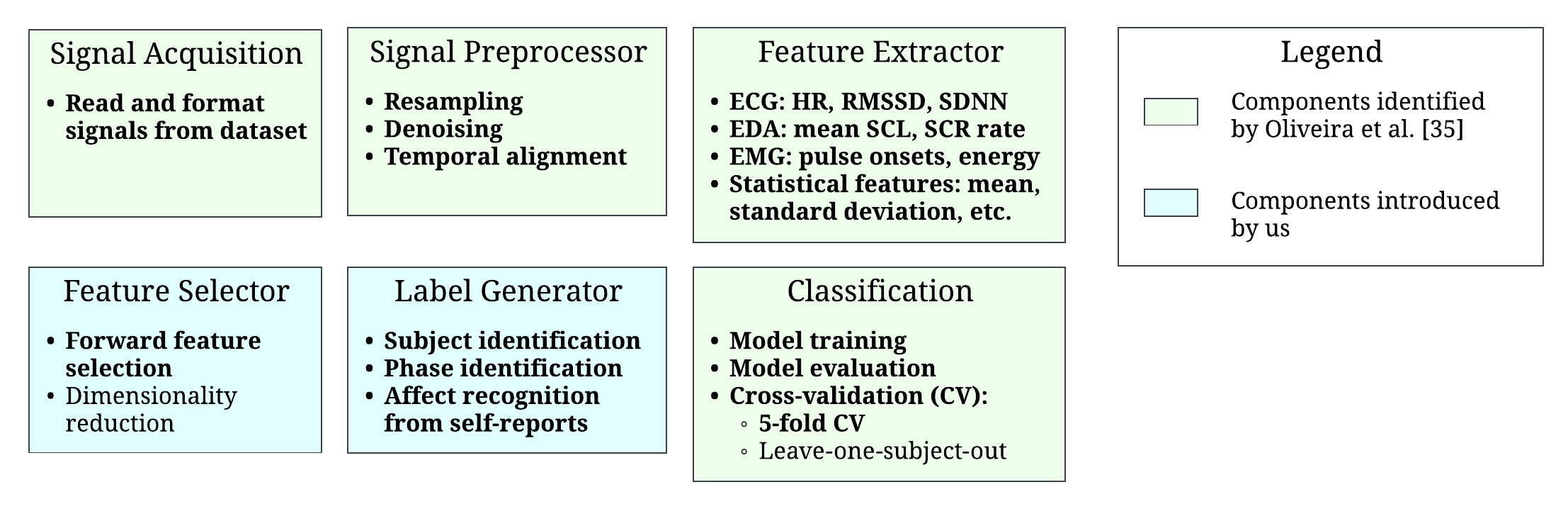}
    \caption{Overview of \toolname~ components. We extend the components identified by \cite{metafera} with Feature Selection and Label Generation, which are useful and often necessary steps for affect recognition across domains. Bolded points indicate pre-implemented behaviors.}
  \label{fig:components}
\end{figure*}

\begin{figure}[!tbp]
    \includegraphics[width=9cm]{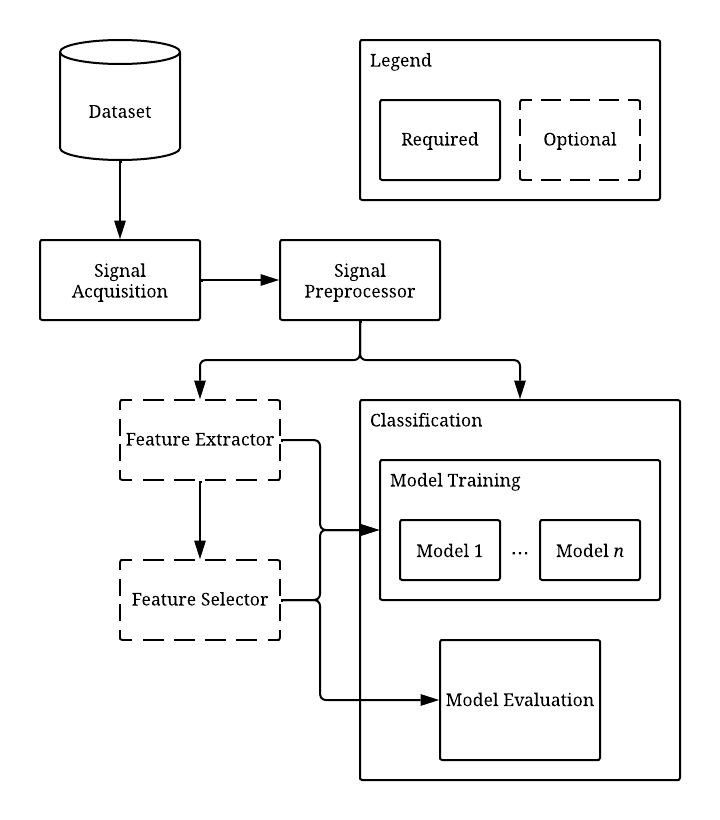}
    \caption{Example instantiation of \toolname. Solid outlines indicate required components, and dashed outlines indicate optional components.}
    \label{fig:example}
\end{figure}

\textbf{Modularity}: \toolname~ adopts a modular architecture, designed based on the common components identified in previous work. It consists of six classes, which we refer to as \textit{components}: (1) Signal Acquisition, (2) Signal Preprocessor, (3) Feature Extractor, (4) Feature Selector, (5) Label Generator, and (6) Classification. We introduce Feature Selection and Label Generation as new components to the original affective computing pipeline outlined by \cite{metafera}. We provide pre-implemented behaviors for each component, which we describe in bolded points in Figure \ref{fig:components}. Components are independent and can be easily modified, and in the case of the Feature Extractor and Feature Selector, omitted as needed. Users instantiate the necessary components, then organize them in an ordered list such that each consecutive component's predefined input and output types are compatible. The required order of components is illustrated in Figure \ref{fig:example}. This list is then passed to the \code{Pipeline}, which executes each component's functions in order.
    
\textbf{Customizability}: \toolname~ allows users to modify and augment its functionalities as needed at both the component- and method-level. Each component has a default implementation that extends from an \code{abstract base class (ABC)}, but users can choose to implement their own versions of each class as needed. Each predefined component also provides default methods that can be overridden. Listing \ref{list:pipeline} presents a high-level view of a pipeline created using \toolname, adapted from our own implementation to replicate Schmidt et al. and Zhou et al. as described in Section \ref{sec:system_evaluation}. Within each component instantiation, parameters such as \code{signal\_types}, \code{feature\_extraction\_methods}, and \code{models} can be specified by the user. 
    
\textbf{Signal- and Domain Agnostic}: \toolname~ can be easily extended to support a wide range of signal types across different applications, \textit{e.g.}, multi-class emotion recognition, stress detection, and depression detection. Generally, existing work focuses on a single domain, but the common components across domains can be implemented and reused with \toolname. For example, Listing \ref{list:pipeline} is an example pipeline that can be modified to perform various affect recognition tasks on different datasets.

Figure \ref{fig:example} provides an example instantiation of \toolname~ and depicts the flow of data between components of the pipeline. The main effort required on the user's end comes from the implementation of the pipeline, the definition of custom components and methods, and the formatting of datasets to be compatible with the Signal Acquisition component, which we discuss below. 

\subsection{Prerequisites}
\label{subsec:prereq}
The structure of the dataset folder and files must follow a standard format\footnote{The \toolname~ GitHub repository (\url{https://github.com/ANRGUSC/AffectEval}) contains example scripts to reformat datasets to the expected format, \code{apd.py} and \code{wesad.py}.} in order for the framework to process input signals for affect recognition tasks. The expected directory structure for data files is as follows:
\code{
\begin{tabbing}
    \{source\_folder\}/ \\
    \hspace{1em} \= \{subject\_ID\_1\}/ \\
    \> \hspace{1em} \= \{subject\_ID\_1\}\_\{phase\}\_\{modality\}.\{csv\} \\
    \> \> ... \\
    \> \{subject\_ID\_2\}/ \\
    \> ...
\end{tabbing}
}

The \code{subject\_ID} value must be a unique identifier for each subject in the dataset. The \code{phase} describes the different phases of the emotion elicitation experiment; typically, affective computing datasets include rest phases to obtain baseline measurements and phases of exposure to various stimuli and stressors. Finally, \code{modality} represents the signal type contained in that file, \textit{i.e.}, ECG, EDA, or RESP, and these signal types must be supported by \toolname~ (currently, ECG, EDA, EMG, RESP, and TEMP signals). The Comma-Separated-Values (CSV) file format, which contains data in a tabular format, is required. The CSV file must contain 2 headers: (1) \textit{timestamp}, and (2) \textit{modality} (\textit{e.g.}, ECG, EDA, or EMG). The \textit{timestamp} column must contain the timestamps for data collection, and the \textit{modality} column must contain the corresponding measurements, \textit{e.g.}, ECG measurements in millivolts, or EDA measurements in microsiemens.

\subsection{Signal Acquisition}
\label{sec:signal_acq}
The \code{Signal Acquisition} component reads signals from CSV files from a dataset folder and formats the data into \code{pandas DataFrame} objects \cite{pandas}. This component returns a dictionary of lists of \code{pandas DataFrames}, where each key in the dictionary corresponds to a unique subject, and each list contains all the signals for that subject as \code{DataFrames}. The set of supported signals can be easily expanded by defining new signal types in the included metadata file.

\subsection{Signal Preprocessor}
The \code{Signal Preprocessor} performs signal denoising and additional preprocessing steps, such as resampling and interpolation. Default preprocessing methods are provided using the \code{biosppy}, \code{heartpy}, \code{neurokit}, and \code{scipy} Python libraries, but users may pass in custom methods via the class parameter \code{preprocessing\_methods}. Specifically, we provide methods for denoising ECG, EDA, EMG, and RESP signals. These include removing baseline wander and powerline noise from ECG and EDA signals and applying filters to remove unnecessary frequencies from EMG and RESP signals. Each \code{DataFrame} is processed separately. This component maintains the same data structure as the \code{Signal Acquisition} component: a dictionary of lists of \code{DataFrames}.

\subsection{Feature Extractor}
The \code{Feature Extractor} extracts high-level features from input signals using either default methods provided by the \toolname~ framework or user-defined methods passed to the \\\code{feature\_extraction\_methods} class parameter. We implemented methods to extract features from ECG, EDA, EMG, RESP, and TEMP signals; these include heart rate and heart rate variability metrics from ECG signals, tonic and phasic components from EDA signals, and statistical features. By default, features are averaged across the feature time series for each phase, but users may override this functionality via the \code{calculate\_average} class parameter. Finally, feature fusion is performed to combine features extracted from different signals. The set of extracted features is concatenated to form a new \code{DataFrame} with each feature in a separate column. The output of this component is a dictionary in which keys correspond to subjects and values are each subject's corresponding feature set, organized in a \code{DataFrame}.

\begin{figure*}[!htbp]
    \begin{python}[caption={High-level code for implementing a pipeline using \toolname~}, label={list:pipeline}]
    # Define label generator function
    def generate_labels():
        ...
    signal_types = [ECG, EDA, EMG]    # Specify signal types
    signal_acq = SignalAcquisition(signal_types, source_folder)
    # preprocessing_methods is not specified, so default methods will be used
    signal_preprocessor = SignalPreprocessor(resample_rate=250)
    # custom methods are specified via feature_extraction_methods
    feature_extractor = FeatureExtractor(feature_extraction_methods)
    label_generator = LabelGenerator(generate_labels)
    models = {
        "DT": DecisionTreeClassifier(criterion="entropy"),
        "AB": AdaBoostClassifier(n_estimators=100),
        "KNN": KNeighborsClassifier(n_neighbors=9)
    }
    # Classification modes: 0 = training, 1 = testing, 2 = cross-validation
    classifier = Classification(mode=2, models=models)
    pipeline = Pipeline()
    pipeline.generate_nodes_from_layers(
        [signal_acq, signal_preprocessor, feature_extractor, 
        label_generator, estimator_train_val_test]
    )
    out = pipeline.run()
    # Obtain results from pipeline
    y_true = out[1]
    y_preds = out[2]
    \end{python}
\end{figure*}

\subsection{Feature Selector}
The \code{Feature Selector} is an \textit{optional} component that can be inserted after the \code{Feature Extraction} component. The default feature selector provided is the \code{SequentialFeatureSelector} from \code{scikit-learn}, but users may use custom feature selection methods that are compatible with the \code{scikit-learn} feature selection module, \textit{e.g.}, users must define \code{fit()} and \code{get\_feature\_names\_out()} methods. This component automatically identifies categorical features and performs one-hot encoding is automatically performed for categorical features to ensure that these features are compatible with ML models, which require numerical input.

\subsection{Label Generator}
The \code{Label Generator} is an additional component we introduced aimed to improve the ease of implementing an affect recognition system. It creates labels based on the features passed from the \code{Feature Extraction} or \code{Feature Selection} component. \toolname~ provides default functions to generate labels based on subject IDs and phase names obtained from the dataset folder. However, to perform affect recognition from annotations or other labels, users are required to implement their own label generator function and pass it to the \code{Label Generator} via the \code{label\_generation\_method} class parameter. This function must generate a corresponding label for each feature vector. For example, WESAD provides self-report files for each subject, consisting of one self-report per questionnaire, per experimental phase. In our experiments, we implemented a function to extract specific questionnaire responses and generate binary affect labels, which we detail in Section \ref{sec:implementation}.

\subsection{Classification}
The Classification component enables the training and testing of affective computing models. Like the Feature Selection component, this component supports any custom methods compatible with \code{scikit-learn}: classifiers must have \code{fit()} and \code{predict()} methods defined, and cross-validation methods must be compatible with the \code{scikit-learn} \code{cross\_val\_score()} method. The default classifier provided is \code{scikit-learn's} Support Vector Machine (SVM) \cite{svm}. Users may change the model, cross-validation method, and execution mode (training, testing, or cross-validation) via class parameters. The output of this component is the final output of the pipeline, returning a list of fitted models, ground-truth labels, and model predictions for training and cross-validation modes. 

%% file: sections/4_system_evaluation.tex
\section{System Evaluation}
\label{sec:system_evaluation}
To demonstrate the validity and reusability of \toolname, we build pipelines to replicate various affect classification tasks originally conducted by Schmidt et al. \cite{schmidt2018} and Zhou et al. \cite{zhou2023} on the following datasets: the Wearable Stress and Affect Dataset (WESAD) \cite{schmidt2018} and the Anxiety Phases Dataset (APD) \cite{senaratne2021}. These datasets were chosen because they are often used in affective computing studies, providing many points of comparison for our framework performance. We reproduce the results of Schmidt et al. \cite{schmidt2018} and Zhou et al. \cite{zhou2023} using 90\% and 89\% fewer lines of code, respectively. We also demonstrate \toolname's modularity and customizability by reusing the same pipeline structure across a wide variety of signals, features, and machine learning models. 

\subsection{Dataset Overview}
\paragraph{Anxiety Phases Dataset (APD)} APD consists of electrocardiogram (ECG), electrodermal activity (EDA), and accelerometer (ACC) recordings from 52 subjects across different lab-controlled emotion elicitation phases. These include rest phases (to obtain baseline measurements), anticipation of an upcoming stressor, exposure to the stressor, recovery post-stressor, and a spoken reflection. The two stressors used in the study were a public speaking task in which subjects were instructed to order three topics by difficulty level, then prepare a 3-minute speech on the most difficult topic, and a bug-box task where subjects were instructed to release a fake bug from a small box (without knowing the bug was fake). Subjects reported their anxiety levels using the SUDS \cite{suds} and LSAS questionnaires \cite{lsas} for each phase of the study. In our reproduction, we found that some subjects did not complete either the speech exposure task or the bug-box exposure task. These subjects were excluded from our experiments. 

\paragraph{Wearable Stress and Affect Dataset (WESAD)} WESAD contains ECG, EDA, ACC, EMG, respiration, and temperature signals from 15 subjects, collected during rest, amusement, stress, and two meditation phases. These signals were collected from both chest- and wrist-based devices, the \textit{RespiBAN Professional} and \textit{Empatica E4}, respectively. The emotion elicitation methods used were 11 humorous video clips and the Trier Social Stress Test (TSST). Subjects also reported their affect levels after each phase using the PANAS \cite{panas}, STAI \cite{STAI}, SAM \cite{sam}, and SSSQ \cite{sssq} questionnaires.

\subsection{\toolname~ Implementation}
\label{sec:implementation}
\begin{figure*}[!htbp]
\centering
\begin{subfigure}{0.45\textwidth}
    \centering
    \includegraphics[height=7.5cm]{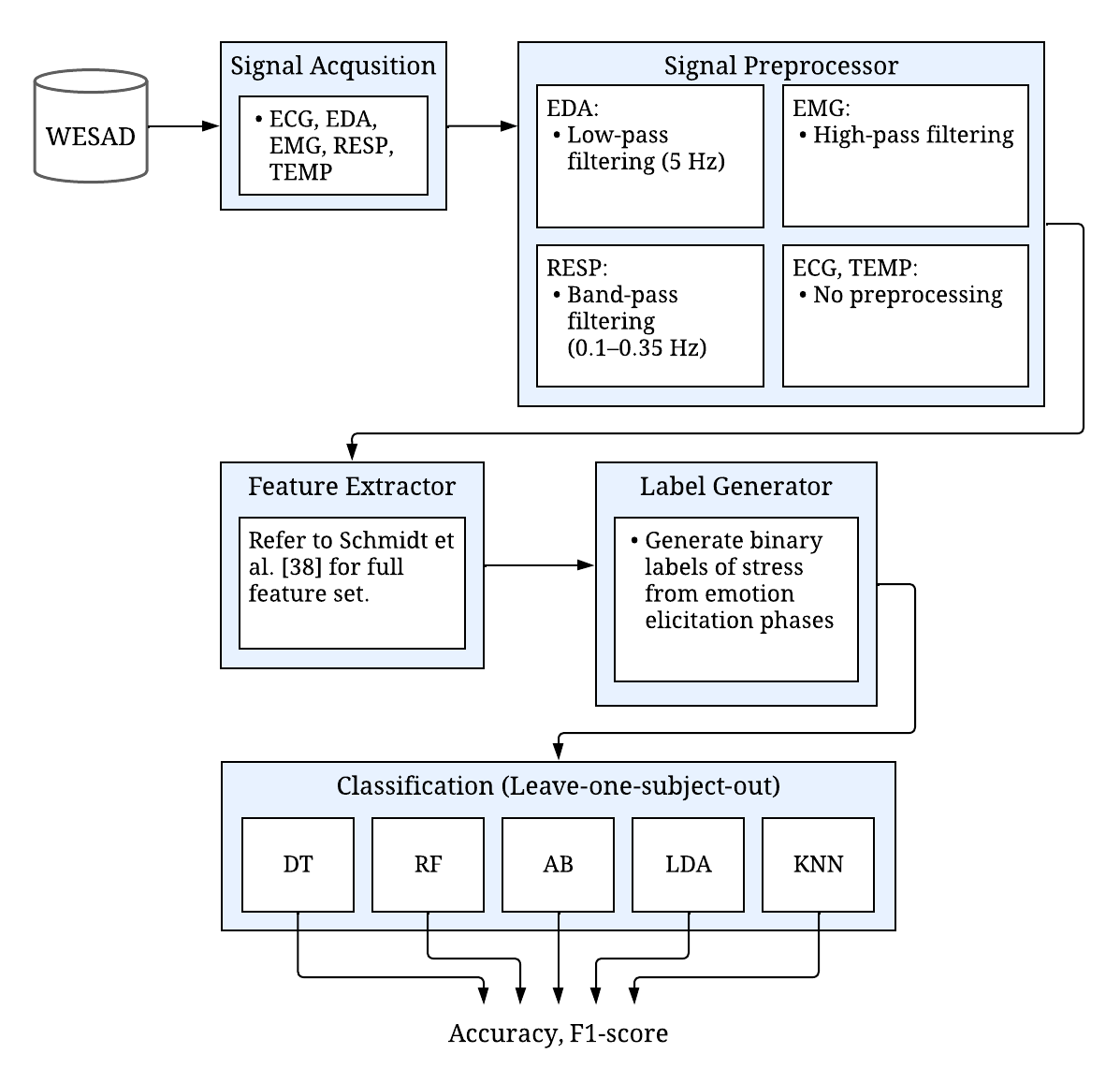} 
    \caption{\toolname~ pipeline to replicate Schmidt et al. \cite{schmidt2018}}
    \label{fig:schmidt}
\end{subfigure}
\begin{subfigure}{0.45\textwidth}
    \centering
    \includegraphics[height=7.5cm]{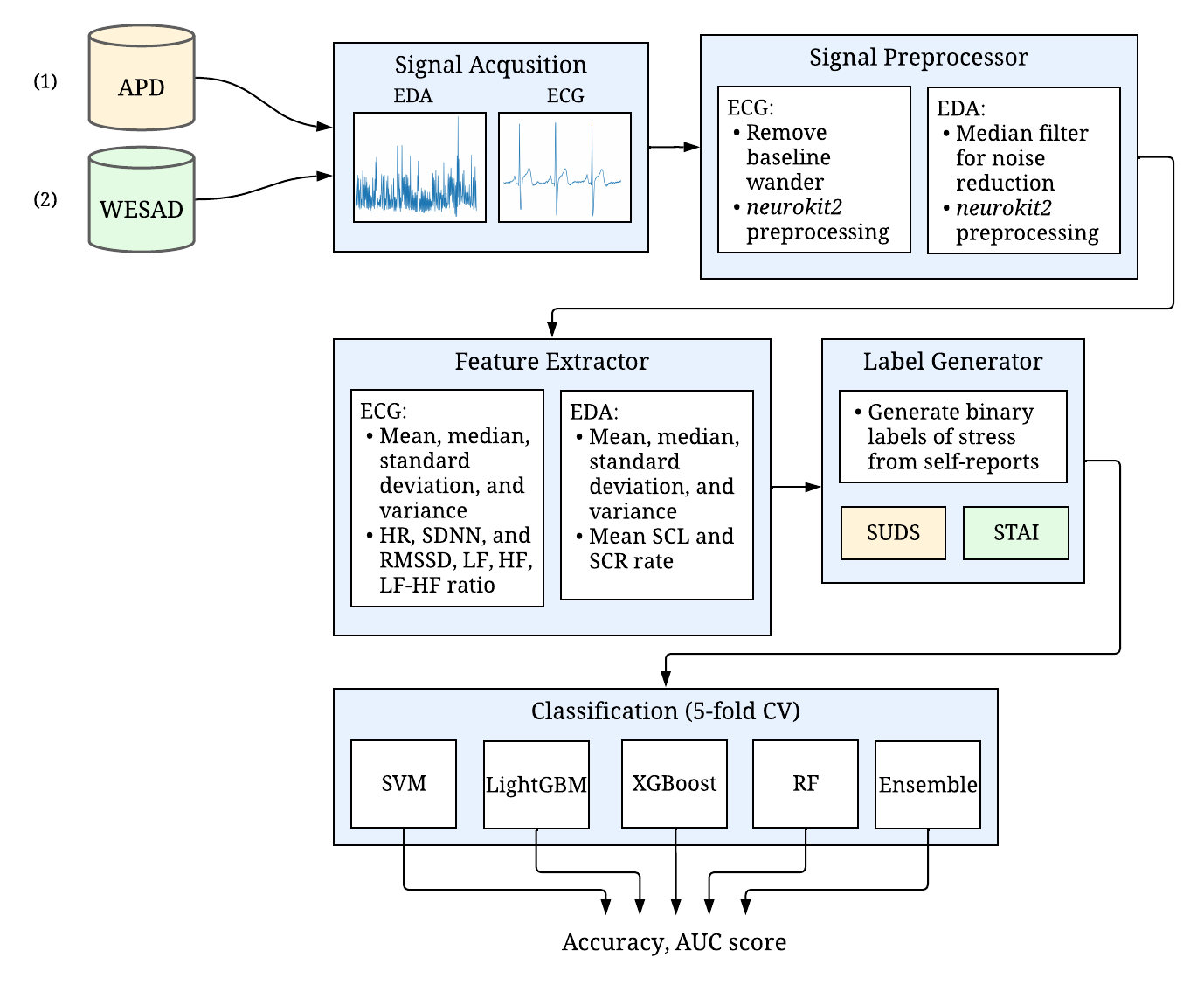}
    \caption{\toolname~ pipeline to replicate Zhou et al. \cite{zhou2023}}
    \label{fig:zhou}
\end{subfigure}
\caption{The workflow of \toolname-based implementation to replicate 3-class affect recognition and binary stress detection on WESAD \cite{schmidt2018} and binary stress detection on APD and WESAD \cite{zhou2023}.}
\label{fig:implementation}
\end{figure*}

We create affective computing pipelines using \toolname~ modules to replicate previous work on binary stress detection on APD and WESAD, which is publicly available in the \toolname~ GitHub repository as Jupyter Notebook files\footnote{Specifically, we reproduce a subset of the findings of Schmidt et al. \cite{schmidt2018} using the physiological modalities from WESAD. We also reproduce the within-corpus results of Zhou et al. \cite{zhou2023} for APD and WESAD.}. In the following sections, we describe the methodologies used and how we implement them with \toolname. Figure \ref{fig:implementation} describes the pipeline components we implemented to replicate these experiments.

\paragraph{Schmidt et al. 2018 Replication}
\label{sec:WESAD}
Schmidt et al. \cite{schmidt2018} performed three-class phase classification \textit{(baseline vs. stress vs. amusement)} and binary stress classification by combining the amusement and baseline conditions to form the \textit{non-stress} class. A total of 16 modality combinations were evaluated across all chest- and wrist-sensor modalities. In this paper, we replicate their experiments on three-class and binary classification tasks using all chest-based physiological modalities - ECG, EDA, EMG, respiration (RESP), and temperature (TEMP) - since this set of modalities achieved the highest accuracy in both tasks. Next, we briefly discuss the preprocessing and feature extraction methods used. 

Physiological signals were segmented using a 60-second window with a 0.25-second overlap. From the raw ECG signals, the following features were extracted: heart rate (HR), mean HR, standard deviation of HR, heart rate variability (HRV), energy in the ultra low, low, high, and ultra-high frequency bands. EDA signals were first filtered with a 5 Hz low-pass filter before statistical features were extracted. Next, signals were decomposed into the tonic and phasic components, also known as the skin conductance level (SCL) and skin conductance response (SCR), respectively. From these components, 9 additional features were extracted. Two sets of features were extracted from EMG signals. First, a high-pass filter was applied to remove the DC component before statistical and frequency-domain features were computed from 5-second windows. Spectral energy was also computed for frequency bands from 0 to 350 Hz. For the second set of features, the raw EMG signal was first filtered with a 50 Hz low-pass filter and segmented in 60-second windows. From each window, peak values and statistical features were calculated. RESP signals were filtered using a band-pass filter with cutoff frequencies of 0.1 and 0.35 Hz. Maxima and minima values and inhalation/exhalation features were extracted. From the raw TEMP signals, statistical features and slope were calculated. The full list of features can be found in Schmidt et al. \cite{schmidt2018}.

The extracted features were concatenated and used as input for classification. The machine learning algorithms used were Decision Tree (DT), Random Forest (RF), AdaBoost (AB), Linear Discriminant Analysis (LDA), and K-nearest neighbors (KNN). Models were evaluated using the leave-one-subject-out (LOSO) cross-validation method, which we implement as a custom cross-validation method using \toolname's \code{Classification} component. To evaluate model performance, accuracy and micro F1-score were calculated.

\paragraph{Zhou et al. 2023 Replication} 
Zhou et al. \cite{zhou2023} performed binary stress classification on APD, WESAD, and the Continuously Annotated Signals of Emotion (CASE) dataset \cite{case} using ECG and EDA signals. Within- and cross-corpus experiments were conducted to evaluate the generalizability of physiological features across stress and high-arousal states. In this paper, we replicate the subset of within-corpus experiments on APD and WESAD. 

ECG and EDA signals were first denoised using \code{biosppy} and \code{neurokit} methods. Segmentation was performed using 60-second sliding windows with a 30-second overlap. 

From ECG signals, the following features were extracted: statistical features (mean, median, standard deviation, and variance), HR, root mean square of successive differences between RR intervals (RMSSD), standard deviation of the inter-beat-interval (SDNN), power in the high (HF) and low frequency (LF) bands, and the ratio of LF to HF. From EDA signals, statistical features, mean SCL, and SCR rate was calculated. 

Stress labels were obtained from subject self-reports. For APD, the SUDS questionnaire responses were used to generate binary labels for stress. The version of SUDS used in APD ranges from 0 to 100, and the median, 50, was used as a fixed value to binarize the subject self-reports. SUDS scores 50 and above were labeled 1, while scores below 50 were labeled 0. For WESAD, the 6-item STAI questionnaire \cite{STAI} responses were used to generate labels. A dynamic threshold was calculated for each subject by taking the average STAI score across all phases. STAI scores greater than or equal to this threshold were labeled 1, while scores less than the threshold were labeled 0. 

The classifiers used were Support Vector Machine, LightGBM, Random Forest, XGBoost, and an ensemble of the previous models. A 5-fold cross-validation method was used, and accuracy and AUC score were reported as model evaluation metrics. 

\input{tables_and_figures/results}

\subsection{Validating \toolname}
We validated \toolname~ by replicating \cite{schmidt2018, zhou2023} to the best of our ability, using the same preprocessing methods, physiological features, labels, and classification models outlined by the authors. To migrate existing work to \toolname, we identified the affective computing components each work used in their experiments: Signal Acquisition, Signal Preprocessing, Feature Extraction, and Classification. Next, we identified the specific preprocessing and feature extraction methods used for each type of signal and implemented them in our \toolname-based pipeline. In our replication of \cite{zhou2023}, we implemented methods to extract 14 features from ECG and EDA signals; for \cite{schmidt2018}, we implemented methods to extract a total of 62 features from WESAD's set of physiological signals. 

Our pipelines achieved the same or better accuracies, AUC scores, and F1-scores across all experiments, which are listed in Table \ref{tab:results}. \textit{By successfully reproducing previous work \cite{schmidt2018, zhou2023}, we show that \toolname~ is effective and comprehensive, providing components and functionalities to perform a wide variety of multimodal affective computing tasks.} In addition, our \toolname~ implementation facilitates future discoveries using APD and WESAD under a fair playground by removing all setup work required. 

While our main goal was to demonstrate the functionality and ease-of-use of \toolname, we were able to outperform previous work in most experiments, as shown by the bolded values. Discrepancies in model performance can most likely be attributed to preprocessing methods selected from different software libraries and differences in random seeding and model parameters that were not specified in previous work. 

\subsection{Reduction in Effort using \toolname~}
\begin{figure}[!htbp]
  \includegraphics[width=0.47\textwidth]{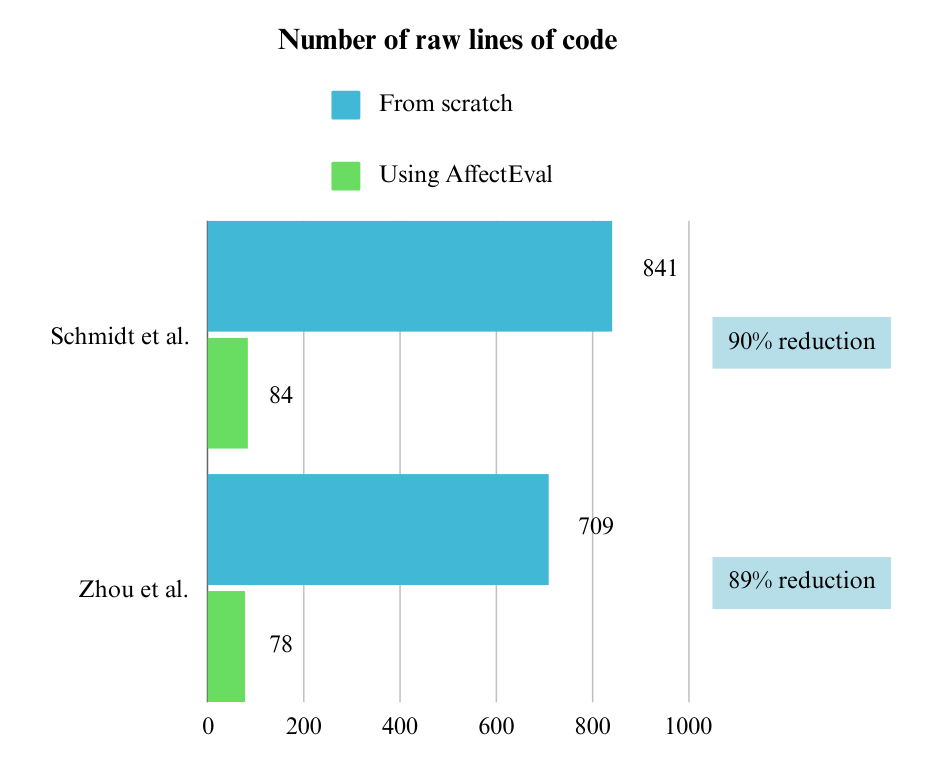}
  \caption{Comparison of \textit{manual effort} required for pipeline implementation.}
  \label{fig:reduction}
\end{figure}

We quantify the reduction in manual effort achieved by using \toolname~ to build our affect recognition pipelines, which we measure through the number of raw lines of code. Since the original code is not publicly available, we estimate the reduction in the number of lines of code we implemented to perform signal preprocessing, feature extraction, and classification. We focus on these components because the remaining tasks of acquiring and formatting datasets and generating labels are necessary whether or not \toolname~ is used. Furthermore, the parameters and custom functions we used to perform signal preprocessing, feature extraction, and classification for these experiments are included as pre-implemented options in \toolname, mitigating the need for users to implement them from scratch. We compare the total lines of code written to the number of lines of code needed to instantiate pipeline components, excluding the implementation of their behaviors. We estimate that \toolname~ reduced the programming effort required to reproduce the experiments in \cite{schmidt2018} by 90\% and in \cite{zhou2023} by 89\%, illustrated in Figure \ref{fig:reduction}. 

%% file: tables_and_figures/results.tex
\begin{table*}[]
\begin{tabular}{cccccc}
\toprule

\multirow{2}{*}{Implementation} & \multirow{2}{*}{\shortstack[c]{Schmidt et al.}}
& \multicolumn{2}{c}{3-Class Classification} & \multicolumn{2}{c}{Binary Stress Classification} \\
\cmidrule(lr){3-4} \cmidrule(lr){5-6}
& & Accuracy (\%) & F1-Score & Accuracy & F1-Score \\ \midrule
\multirow{5}{*}{\rotatebox{90}{Original}}
& DT & 58.62$_{\pm{1.07}}$ & 55.10$_{\pm{0.92}}$ & 84.18$_{\pm{0.20}}$ & 81.29$_{\pm{0.22}}$ \\
& RF & 71.37$_{\pm{0.58}}$ & 64.60$_{\pm{0.54}}$ & 92.01$_{\pm{0.51}}$ & 90.44$_{\pm{0.66}}$ \\
& AB & 80.34$_{\pm{0.43}}$ & 72.51$_{\pm{0.17}}$ & 89.76$_{\pm{0.48}}$ & 87.11$_{\pm{0.57}}$ \\
& LDA & \textbf{79.35} & \textbf{74.43} & \textbf{93.12} & \textbf{91.47} \\
& KNN & 56.14 & 48.70 & 81.05 & 77.27 \\
\midrule
\multirow{5}{*}{\rotatebox{90}{\toolname}} 
& DT & \textbf{81.68$_{\pm{3.25}}$} & \textbf{80.94$_{\pm{4.38}}$} & \textbf{95.48$_{\pm{1.57}}$} & \textbf{93.94$_{\pm{2.01}}$} \\
& RF & \textbf{90.52$_{\pm{2.98}}$} & \textbf{91.1$_{\pm{2.18}}$} & \textbf{95.50$_{\pm{1.07}}$} & \textbf{95.18$_{\pm{1.71}}$} \\
& AB & \textbf{77.60$_{\pm{4.10}}$} & \textbf{76.56$_{\pm{4.13}}$} & \textbf{93.10$_{\pm{0.90}}$} & \textbf{93.07$_{\pm{0.82}}$} \\
& LDA & 69.83$_{\pm{2.61}}$ & 66.5$_{\pm{2.69}}$ & 83.61$_{\pm{3.19}}$ & 82.58$_{\pm{3.80}}$ \\
& KNN & \textbf{76.73$_{\pm{3.73}}$} & \textbf{74.51$_{\pm{4.08}}$} & \textbf{87.29$_{\pm{5.14}}$} & \textbf{86.76$_{\pm{5.75}}$} \\

\midrule \midrule

\multirow{2}{*}{Implementation} & \multirow{2}{*}{\shortstack[c]{Zhou et al.}}
& \multicolumn{2}{c}{APD} & \multicolumn{2}{c}{WESAD} \\
\cmidrule(lr){3-4} \cmidrule(lr){5-6}
& & Accuracy (\%) & AUC Score & Accuracy (\%) & AUC Score \\ \midrule
\multirow{5}{*}{\rotatebox{90}{Original}}
& SVM & 54.9 & 49.8 & \textbf{86.0} & \textbf{72.0} \\
& LightGBM & 54.7 & 52.9 & 83.6 & 72.0 \\
& XGBoost & 55.3 & 54.0 & 84.5 & 73.5 \\
& RF & 55.3 & 54.0 & 84.2 & 77.5 \\
& Ensemble & 56.1 & 52.8 & \textbf{99.0} & \textbf{96.9} \\
\midrule
\multirow{5}{*}{\rotatebox{90}{\toolname}} 
& SVM & \textbf{79.17$_{\pm{0.36}}$} & \textbf{59.33$_{\pm{4.77}}$} & 57.16$_{\pm{1.25}}$ & 61.46$_{\pm{4.91}}$ \\
& LightGBM & \textbf{88.57$_{\pm{1.60}}$} & \textbf{89.91$_{\pm{2.22}}$} & \textbf{96.54$_{\pm{1.97}}$} & \textbf{99.58$_{\pm{0.43}}$} \\
& XGBoost & \textbf{87.78$_{\pm{1.25}}$} & \textbf{89.76$_{\pm{2.30}}$} & \textbf{95.16$_{\pm{2.06}}$} & \textbf{98.46$_{\pm{1.60}}$} \\
& RF & \textbf{86.99$_{\pm{1.25}}$} & \textbf{91.19$_{\pm{1.19}}$} & \textbf{95.28$_{\pm{2.07}}$} & \textbf{99.25$_{\pm{0.57}}$} \\
& Ensemble & \textbf{86.29} & \textbf{67.52} & 86.29 & 67.53 \\

\bottomrule

\end{tabular}
\caption{A comparison of \toolname's performance to Schmidt et al. \cite{schmidt2018} and Zhou et al.'s \cite{zhou2023} implementations. Abbreviations: DT = Decision Tree, RF = Random Forest, AB = AdaBoost DT, LDA = Linear Discriminant Analysis, KNN =
K-nearest neighbors, SVM = Support Vector Machine. Ensemble refers to an equally weighted average ensemble of SVM, LightGBM, XGBoost, and RF. Differences in model performance can most likely be attributed to the use of preprocessing and feature extraction methods from different libraries, \textit{e.g.}, biosppy or neurokit, and differences in model parameters not specified by previous work.}
\label{tab:results}
\end{table*}

%% file: sections/5_discussion.tex
\section{Discussion}
\label{sec:discussion}
Our successful reproduction of previous work highlights the contributions of \toolname~ as a framework for creating affective computing pipelines. We showed that \toolname~ supports multimodal, multi-domain applications and quantify the reduction in manual work that can be achieved by using \toolname. This section compares \toolname~ to other existing affective computing frameworks and illustrates the role of modularity and customizability in making it a more comprehensive and flexible framework.

\subsection{Framework Comparison}
Existing affective computing frameworks, metaFERA \cite{metafera} and AffectToolbox \cite{affecttoolbox}, were not evaluated by replicating previous work, or across a variety of affective computing tasks. As a result, there is no point of comparison for their ease of use, or demonstration of multimodal and multi-domain capabilities. 

To validate metaFERA, authors created a framework for EDA-based binary affect detection of high vs. low arousal. However, it is unclear how much less manual work the metaFERA-based pipeline requires, compared to implementing such an EDA-based framework from scratch. Additionally, metaFERA \textit{does not provide pre-implemented behavior}, requiring users to define component behaviors. Therefore, it is likely that metaFERA has a higher manual overhead than \toolname. While metaFERA is modular and customizable, its effectiveness in reducing manual work and creating multi-domain applications is unclear. 

The main goal of AffectToolbox is to foster a more collaborative environment in affective computing by removing the need for any programming knowledge. It provides a graphical user interface (GUI) to build affective computing pipelines for pleasure, arousal, and dominance classification from audiovisual data streams. While AffectToolbox effectively reduces the manual work involved in creating such pipelines, it is less comprehensive than \toolname~ due to its limited support for multimodal signals and affect recognition capabilities. 

\subsection{Modularity and Customizability}
In comparison, the modularity of \toolname~ allows for the reuse of pipeline structure across multimodal, multi-domain affective computing applications, while the individual components can be customized to perform application-specific tasks. Our replication of the experiments conducted by Schmidt et al. and Zhou et al. reuses the \code{Signal Preprocessing}, \code{Feature Extraction}, \code{Label Generation}, and \code{Classification} components. We customized these components to use different signal preprocessing techniques, feature extraction methods, affect labels, and classification models across experiments by changing the parameters of each component. The ease of switching out components and changing their behaviors also enabled us to set up a fair playground to compare the performance of several classification models. Our successful reproduction of Schmidt et al. and Zhou et al.'s experiments demonstrates that \toolname~ effectively mitigates the challenges of high manual effort and redundant code between pipelines, while being flexible, supporting a wide variety of signals and affective computing tasks.

\subsection{Limitations}
Through the development and validation of \toolname, we identify a few remaining limitations and suggest directions for future improvements.

\toolname~ primarily supports the use of time-series physiological signals and lacks pre-defined functionalities for image, text, and audio data. Due to their increasing popularity and high performance in affect detection, we prioritized the support of time-series physiological signals. However, facial expression recognition, body gestures, audio, and textual data are non-invasive data streams that are also commonly used in affective computing \cite{wang2022}. As mentioned in Section \ref{sec:signal_acq}, \toolname's capabilities can be easily expanded to support other types of signals.

\toolname~ does not support real-time emotion recognition or distributed computing. This limits the applicability of \toolname-based pipelines in online emotion recognition tasks. Additionally, components such as signal preprocessing, feature extraction, and model training cannot be parallelized, limiting the computing and time efficiency of \toolname-based pipelines.

These limitations can be addressed by extending \toolname's capabilities and list of pre-implemented behaviors. Specifically, we plan to add additional preprocessing methods to support audiovisual and textual data. In addition, a cloud-based or client-server wrapper can be created, enabling \toolname~ to support distributed and real-time processing capabilities. Finally, \toolname~ can be enhanced with a GUI, similar to \cite{affecttoolbox}, to further improve its usability and inclusivity by removing the need for programming knowledge.

%% file: sections/6_conclusion.tex
\section{Conclusion}
\label{sec:conclusion}
This paper presented \toolname, a comprehensive framework to support the development of affective computing pipelines. It addresses key shortcomings of software libraries and tools currently used in affective computing applications, reducing the effort required to set up end-to-end pipelines. \toolname~ was validated on previous work, exemplifying the reduction in manual work required while achieving the same or higher affect recognition performance. \toolname~ is publicly released, with the goal of facilitating future affective computing research and creating an open-source repository for the community to contribute to.